\pdfoutput=1
\documentclass[11pt]{article}

\usepackage{ACL2023}

\usepackage{times}
\usepackage{latexsym}
\usepackage{algorithmic}
\usepackage{algorithm}
\usepackage{array}
\usepackage{textcomp}
\usepackage{latexsym}
\usepackage{amsfonts,amssymb}
\usepackage{amsmath}
\usepackage{graphicx}
\usepackage{graphics}
\usepackage{hyperref}
\usepackage{booktabs}
\usepackage{multirow}
\usepackage{pdfpages}
\usepackage[T1]{fontenc}

\usepackage[utf8]{inputenc}

\usepackage{microtype}

\usepackage{inconsolata}

\title{Constructing Word-Context-Coupled Space Aligned with Associative Knowledge Relations for Interpretable Language Modeling}

\author{Fanyu Wang \and Zhenping Xie\footnotemark[1]{} \\
        School of Artificial Intelligence and Computer Science, Jiangnan University, China \\ \texttt{fanyu\_wang@stu.jiangnan.edu.cn} \\ \texttt{xiezp@jiangnan.edu.cn} \\}

\begin{document}
\maketitle

\renewcommand{\thefootnote}{\fnsymbol{footnote}}
\footnotetext[1]{Corresponding author}
\renewcommand{\thefootnote}{\arabic{footnote}}

\begin{abstract}
As the foundation of current natural language processing methods, pre-trained language model has achieved excellent performance. However, the black-box structure of the deep neural network in pre-trained language models seriously limits the interpretability of the language modeling process. After revisiting the coupled requirement of deep neural representation and semantics logic of language modeling, a Word-Context-Coupled Space (W2CSpace) is proposed by introducing the alignment processing between uninterpretable neural representation and interpretable statistical logic. Moreover, a clustering process is also designed to connect the word- and context-level semantics. Specifically, an associative knowledge network (AKN), considered interpretable statistical logic, is introduced in the alignment process for word-level semantics. Furthermore, the context-relative distance is employed as the semantic feature for the downstream classifier, which is greatly different from the current uninterpretable semantic representations of pre-trained models. Our experiments for performance evaluation and interpretable analysis are executed on several types of datasets, including SIGHAN, Weibo, and ChnSenti. Wherein a novel evaluation strategy for the interpretability of machine learning models is first proposed. According to the experimental results, our language model can achieve better performance and highly credible interpretable ability compared to related state-of-the-art methods.\footnote{https://github.com/ColeGroup/W2CSpace}

\end{abstract}

\section{Introduction}

Machine learning has recently been democratized in various domains, such as search engines, conversational systems, and autonomous driving\cite{gao2018conversational, Grigorescu2020autonomous, wang2022survey}. However, despite AI technologies significantly facilitating industrial processes and improving work experiences, the uninterpretable logic of machines leads to distrust, which hinders further development of AI. Explainable Artificial intelligence (XAI), proposed to bridge the block between humans and machines, has been increasingly attracting attention recently, where "explanation" is described as abductive inference and transferring knowledge \cite{josephson1996abductive, Miller2019explanation}. Calling for explaining and understanding the machine learning process, researchers aim to interpret the methods for system verification, compliance with legislation, and technology improvement. 
% Besides, interpretable AI not only increases the working efficiency but also builds the trustworthiness of the machines, especially in the domains related to health, safety, and economy \cite{Li2022interpretable}.

Computational linguistics, which serves as the theoretical foundation for NLP, aims to promote communication between humans and machines \cite{khan2016survey}. However, during the recent development, the uninterpretable NLP methods have raised concerns. The decreased transparency and increased parameter complexity adversely affect the model explainability and controllability, even if the performance of the language models has significantly improved, such as BERTs \cite{devlin-etal-2019-bert, liu2019roberta, Clark2020ELECTRA}, GPTs \cite{radford2018gpt, radford2019gpt2, Brown2020gpt3} and so on. The performance advantage of the black-box methods is attractive while the researchers investigate interpretable algorithms. Therefore, the existing works mainly focus on (1) explaining the black-box methods and (2) using interpretable models \cite{Ribeiro2016why, Li2022interpretable}.

In order to explain over-parameterized language models, the model-agnostic analysis is investigated on recent deep neural methods. Without modifying the black-box models, the researchers analyze the immediate feature of the neural layers, attention distribution, and so on \cite{clark-etal-2019-bert, vig-2019-multiscale, rogers-etal-2020-primer}. Quantitative experiments and visual analysis is able to partially reveal the behaviors of the key components and the overall response of the methods to certain patterns \cite{hewitt-manning-2019-structural, kovaleva-etal-2019-revealing}. However, without any interpretable optimization of the model, the analysis is unable to provide enough detail for understanding \cite{Rudin2019Stop}, which indicates \emph{a completely faithful explanation from black-box components or deep neural methods is impossible}. 

Different from model-agnostic analysis, methods that integrate interpretable algorithms enable more comprehensive interpretability. Two different types of structures are adopted in these approaches, including (1) a black-box backbone with an interpretable bypass for implicit informing and (2) a transparent backbone with an interpretable algorithm for direct interpreting \cite{Katharina2021explainable}. For implicitly informed methods, the introduced interpretable knowledge regulates the immediate feature or embedding \cite{liu-etal-2019-knowledge, Rybakov2020learning}. Within the interpretable bypass, the performance of the backbone is maximally preserved, which is the main reason these structures are often opted for over-parameterized models \cite{jang-etal-2021-kw, Chen2020Concept}. However, \emph{the integrated knowledge is unable to decisively change the structure of the backbones for a transparent decision process}, which adversely affects the generalization ability of the approaches and limits the application of the methods to specific tasks. In contrast, approaches with interpretable backbones exhibit a more integrated relationship between the components, enabling better explanations than the above two types of approaches. The interpretable algorithms serve as word embedding, immediate feature, or the classifier to realize transparency decision process \cite{onoe-durrett-2020-interpretable, lee-etal-2022-toward, kaneko-etal-2022-interpretability}. But \emph{the performance of existing interpretable models remains incomparable to the most advanced language models}.

\begin{figure*}[!t]
    \begin{center}
        \includegraphics[width=5in]{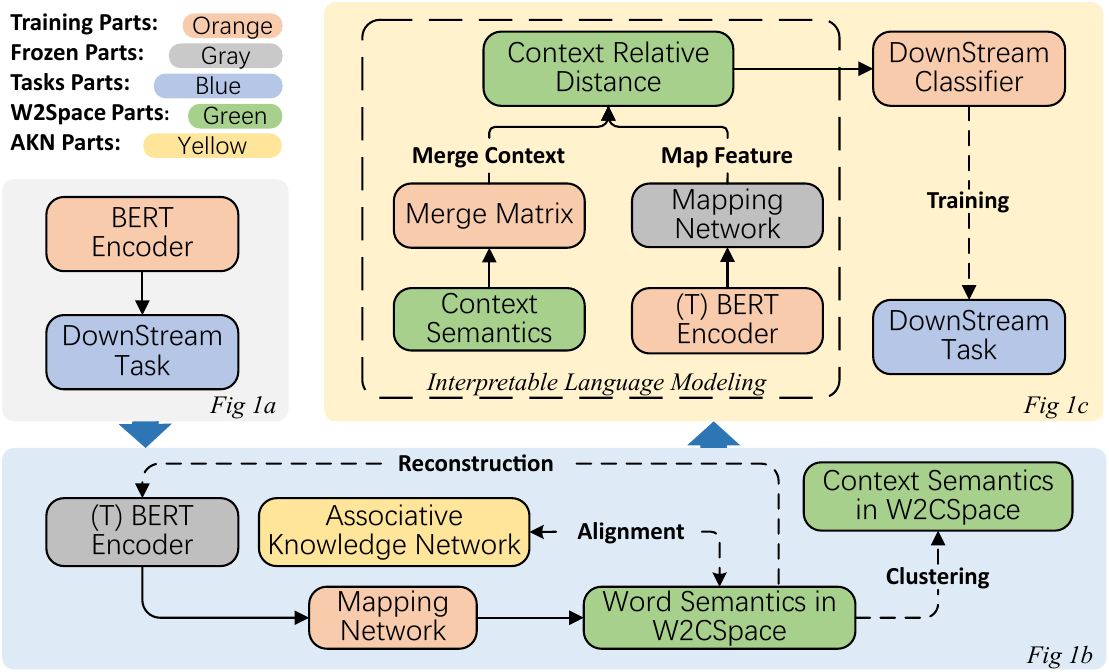}
    \end{center}
    \caption{Overview of the method's architecture. The subfigs exhibit the training process for our interpretable language method. (1a) The BERT encoder is firstly fine-tuned under the downstream task. We map the feature $\boldsymbol{F}_B$ from fine-tuned BERT encoder to the word elements $\boldsymbol{C}$ in W2CSpace ("T" refer to "trained"); (1b) After the training of the mapping network with a reconstruction and alignment task, the word semantics are clustered to high-dimensional context semantics $\boldsymbol{X}$; (1c) During interpretable language modeling process, the context semantics in W2CSpace is gradually optimized by multiplying the merge matrix $\boldsymbol{M}_\mathrm{M}$. We compute the context-relative distance of the mapped feature $\boldsymbol{C}$ within the optimized context $\Bar{\boldsymbol{X}}$ for the downstream task.}
    \label{fig:1}
\end{figure*}

In this work, we address the aforementioned obstacles by developing a novel interpretable language modeling method by constructing a \textbf{W}ord-\textbf{C}ontext-\textbf{C}oupled \textbf{Space} (W2CSpace) aligned with statistical knowledge\footnote{Chinese characters and English words are typically considered at the same processing level in pre-trained language models, so we refer to Chinese characters as "Chinese word" in this paper to avoid ambiguity.}, which enables (1) effective interpretation of BERT representations \cite{devlin-etal-2019-bert} by introducing the interpretable statistical logic, (2) reasonable context abstraction with the coupled word-level semantics, and (3) interpretable modeling for the given text with the context-relative distance. W2CSpace serves as the key component in the backbone of our language modeling method, which realizes a decisive transparency increasing compared with the model-agnostic and implicitly informed methods. The structure of our method is illustrated in Figure \ref{fig:1}. Specifically, our main contributions can be summarized as follows:
\begin{itemize}
    \item Word-level semantics in W2CSpace is originated from BERT immediate representation with the help of a mapping network, preserving the language modeling ability of the deep neural methods (Section \ref{mapping}).
    \item An associative matrix sampled from associative knowledge network (AKN, \citealp{Li2022associative}) is introduced for aligning with the semantic distances (Section \ref{akn} and \ref{mapping training}).
    \item Based on the linguistic concept, the contexts are abstracted using $k$-means clustering on neighboring word elements. (Section \ref{context}).
    \item The context-relative distance, computed between the input text and the context clusters in W2CSpace, serves as the semantic feature to describe the text semantics (Section \ref{crd}).
    \item The experiments on different NLP tasks demonstrate the effectiveness of W2CSpace. Additionally, an interpretable analysis is designed to verify the interpretability of the our method (Section \ref{experiment}).
\end{itemize}

\section{Methodology}
\subsection{Initialization of W2CSpace}

Since current researchers opt for high-dimensional representations in their methods, it is widely believed that over-parameterization advances language modeling performance. With respect to the standard language modeling process, the words in the given text are modeled based on their meaning under different contexts. Regardless of the attributes of the words themselves, greater representation dimensions enable better performance in distinguishing the words with similar semantics. However, different from the deep neural methods, the linguistic attributes of the text, such as co-occurrence rules, word shape, and so on, serve as the basis for NLP tasks, which fit the understanding and deduction processes of humans.

In order to unify the high-dimensional representations and interpretable statistical knowledge, we design a mapping network to transfer the semantic representation from BERT encoder to low-dimensional elements in W2CSpace and introduce a statistical alignment with AKN during the training of the mapping network. Within the above processing, the mapped elements are distributed in W2CSpace according to their corresponding word-level semantics.

\subsubsection{Representation Mapping from BERT}
\label{mapping}
The mapping network is a neural network with a backbone of a convolution network, which enables dimension reduction process. The mapped elements in smaller dimensions are regarded as the coordinates of the word-level semantics in W2CSpace. Besides, the introduction of the convolution network is able preserve semantic information from BERT for maintain the performance advantages of pre-trained models. For BERT representations $\boldsymbol{F}_B$ of given sentence $\mathrm{S}=\{x_1, x_2,\dots,x_d\}$, the corresponding elements $\boldsymbol{C}=\{c_1, c_2,\dots,c_d\}$ in W2CSpace are obtained according to:
\begin{equation}
    \boldsymbol{C}=\mathrm{Tanh}\{\mathrm{LN}[\mathrm{Convs}(\boldsymbol{F}_B)+\mathrm{Res}(\boldsymbol{F}_B)]\}
    \label{eq:1}
\end{equation}
where $\boldsymbol{F}_B \in {\mathbb{R}^{n\times h}}$ and $\boldsymbol{C} \in {\mathbb{R}^{n\times k}}$, $h$ is the hidden sizes of BERT encoder and $k$ is the coordinate size of W2CSpace. $\mathrm{Tanh}(\cdot)$ and $\mathrm{LN}(\cdot)$ are Tanh and layer normalization operations. A convolution network $\mathrm{Convs}(\cdot)$ with different filter sizes added a residual connection $\mathrm{Res}(\cdot)$ is used in the mapping network.

\subsubsection{Statistical Alignment with AKN}
\label{akn}
AKN, a statistical network based on phrase co-occurrence, is introduced to sample to an associative matrix reflected the associative relations among the given sentence, which is also adopted in previous work (AxBERT, \citealp{wang2023axbert}). Within the original AKN is conducted on phrase-level, we modify AKN to word-level to fit the processing of BERT and opt for the construction and sampling methods of the AKN ($\boldsymbol{A}$) and associative matrix ($\boldsymbol{M}_{S}$) similar to AxBERT as:
\begin{equation}
    \boldsymbol{A}_{i,j}=\prod_{\mathrm{sent}}\mathrm{SR}\sum_{\mathrm{sent}} \frac{1}{distance_{\langle i,j \rangle}}
    \label{eq:2}
\end{equation}
\begin{equation}
    {\boldsymbol{M}_{S}}_{i,j}=\sigma\frac{\boldsymbol{A}_{i,j}}{\mathrm{Avg}({\dot{\boldsymbol{A}}_{i:}})}-0.5
    \label{eq:3}
\end{equation}
where $\boldsymbol{A} \in {\mathbb{R}^{v\times v}}$, $\boldsymbol{M}_{S} \in {\mathbb{R}^{d\times d}}$, $v$ is the length of the word list, $d$ is the length of the given sentence, and $\sigma(\cdot)$ and $\mathrm{Avg}(\cdot)$ are functions of sigmoid and average. For the word pair $\langle i,j \rangle$, $distance_{\langle i,j \rangle}=|i-j|$ is the word-distance between $i$-th and $j$-th word in sentence. $\dot{\boldsymbol{A}}_{i:}$ is the association score of $i$-th word under current sentence, and ${\boldsymbol{M}_{S}}_{i:}$ is the $i$-th row of $\boldsymbol{M}_{S}$. \footnote{The same shrink rate $\mathrm{SR}=0.95$ as AxBERT.}

Since we compute the cosine distances matrix in the given sentence, the associative matrix is aligned with the word-level distance matrix to integrate the statistical logic into W2CSpace. We introduce a mean square indicator $I_{MS}$ to indicate the alignment result. Specifically, for the word pair $\langle i,j \rangle$, the indicator ${I_{MS}}_{i,j}$ is defined as:
\begin{equation}
    {I_{MS}}_{i,j}=\mathrm{MnSqr}[\mathrm{CosDis}(c_i,c_j),{{\boldsymbol{M}}_{S}}_{i,j}]
    \label{eq:4}
\end{equation}
where ${I_{MS}} \in {\mathbb{R}^{n\times n}}$ and $n$ is the length of the given sentence. $\mathrm{MnSqr}(\cdot)$ and $\mathrm{CosDis}(\cdot,\cdot)$ are the mean square and cosine distance functions.

\subsubsection{Training of Mapping Network}
\label{mapping training}
The objective $L_M$ of the mapping network is composed of $L_{MS}$ and $L_{Rec}$, which correspond to the mean square error loss and the reconstruction loss. With respect to the statistical alignment process, the mapping network is trained under the alignment objective $L_M$. Besides, we introduce a reconstruction loss by reversing the structure of mapping network to reconstruct the representation of BERT immediate feature $\boldsymbol{F}_B$. The objective of the mapping network is calculated according to:
\begin{equation}
    L_M=L_{MS}+L_{Rec}
    \label{eq:5}
\end{equation}
\begin{equation}
    L_{MS}=\mathrm{Mean}(I_{MS})
    \label{eq:6}
\end{equation}
\begin{equation}
    L_{Rec}=\mathrm{MAE}(\boldsymbol{C}-\boldsymbol{F}_B)
    \label{eq:7}
\end{equation}
%\begin{equation}
%    \boldsymbol{C}_{vs}=\mathrm{Mid}(\sum ^{v} _{i=0} \sum ^{s} _{j=0} c_{i})
 %   \label{eq:8}
%\end{equation}
where $\mathrm{Mean}(\cdot)$ is the average function, $\mathrm{MAE}(\cdot)$ is the mean absolute error operation \cite{Choi_2018_CVPR}. The introduction of reconstruction loss aims to guarantee that the mapped word elements preserve the semantics of BERT representations.

\subsection{Abstraction of Context-level semantics}

\label{2.2}

Humans are able to recognize emotion from language, action, and so on \cite{Barrett2007}. Specifically, in linguistics, humans recognize emotion with the context in the given sentences. However, humans are able to feel the emotion rather than explicitly describe it, because that context is an abstract concept in linguistics and is hard to simply quantify.

While the sentence is composed of words, the corresponding context is established from the word semantics, which can be realized in W2CSpace. Therefore, we employ the $k$-means clustering on the word-level semantics to abstract the context semantics. The context is able to be extracted based on the common semantics among the words located adjacently in W2CSpace.

\subsubsection{Context Clustering Based on Word Semantics}
\label{context}
With the help of the mapping network and the statistical alignment, the word elements are reasonably distributed in W2CSpace according to their semantics, where the neighbors in W2CSpace refer to similar semantics. $k$-means clustering is an algorithm based on the distances \cite{Hartigan1979}, which is introduced to abstract the word semantics to $k$ classes according to their semantics (distance). The clustering process is defined as:
\begin{equation}
    \boldsymbol{X}\{x_i|\mathrm{Context}(x_i)\}=\mathrm{KM}_{\mathrm{CosDis}}(c_1,c_2,\dots,c_n)
    \label{eq:9}
\end{equation}
where $x_i$ is the $i$-th classes of context and $\mathrm{KM}_{\mathrm{CosDis}}(\cdot)$ is the $k$-means clustering algorithm based on cosine distance. The word semantics in W2CSpace is clustered into $k$ classes, which represents the process of abstraction of context. 

\subsubsection{Reasonable Cluster Merging for Context Clustering}

When the clustering is executed, especially in $k$-means algorithm, the appropriate $k$ number is hard to determine \cite{learning2003hamerly}. As for the clustering process for context, the number $k$ additionally represents the types of the contexts in W2CSpace. A small number of $k$ possibly decreases the performance of language modeling with a rough context environment, but the large number is contrary to the logic of humans as humans cannot precisely distinguish the detailed emotion behind the text as a machine does. Besides, the large number of $k$ increase the time-costing of the language modeling process. Choosing the right number $k$ relate to the reasonability of the context subspace.

Serving as the part of W2CSpace, the context clusters are used for language modeling. Therefore, we introduce a merge matrix to the top of the clustering results, which is optimized under the downstream task. With the guidance of the downstream tasks, the merge matrix dynamically adjusts context semantics for the inter-communication between different context clusters, which reflects a gradual clustering process and realizes a reasonable context clusters for the downstream tasks. The clustering is able to be balanced with the computation process of the merging is defined as:
\begin{equation}
    \Bar{\boldsymbol{X}}\{\Bar{x}_i|\mathrm{Context}(\Bar{x}_i)\}=\boldsymbol{M}_\mathrm{M} \times [x_1, x_2, \dots, x_k]
    \label{eq:10}
\end{equation}
where $\Bar{x}_i$ indicates the context semantics after merging, $\boldsymbol{M}_\mathrm{Merge}$ is the merge matrix and $\boldsymbol{M}_\mathrm{Merge} \in {\mathbb{R}^{n\times k}}$. $n$ is the size of coordinates in W2CSpace and $k$ is the presetting number of the clustering. \footnote{However, the number $k$ will influence the performance of language modeling, which is discussed in Section \ref{main result}.}

\subsection{Interpretable Language Modeling via W2CSpace}

As the standard methodology for language modeling in deep neural methods, the semantic representation is gradually modeled through containing neural networks, which is a simulation of the neural processing of the brain. However, the structural simulation is unable to realize the interpretable on logical level. The decision process through neural networks still remains in the black box.

From the perspective of humans, emotion recognition is significant in daily \cite{Barrett2011}, which is also an important ability for machines to interact with humans \cite{Kosti_2017_CVPR}. By simulating the recognition process of humans, we introduce a context-relative distance computed between the given text and the contexts in W2CSpace, which enables the interpretable language modeling process with the cooperation of the semantics on word- and context-level.

\subsubsection{Computation of Context-Relative Distance}
\label{crd}
The context-relative distance based on the cosine distance is also adopted in Formula \ref{eq:4}. Compared with the euclidean distance, the equation of the cosine distance is more efficient in time-costing and storage. The context-relative distance $\boldsymbol{D}$ is computed according to:
\begin{equation}
    \boldsymbol{D}=\mathrm{CosDis}(\Bar{\boldsymbol{X}},\boldsymbol{C})
    \label{eq:11}
\end{equation}
where $\Bar{\boldsymbol{X}}$ is the context clusters, $\boldsymbol{C}$ is the mapped word elements from BERT encoder, $\boldsymbol{D}$ is the context-relative distance and $\boldsymbol{D} \in {\mathbb{R}^{d\times k}}$. $n$ is the word length of the given text and $k$ is the number of the context clusters. 

\subsubsection{Training of Interpretable Language Modeling Method}

The context-relative distance is able to directly connect with the downstream classifier, which is similar to the traditional encoding structure. The interpretable language modeling component is regarded as the standard BERT-based encoder for downstream tasks, where the standard objectives in transformer package\footnote{https://pytorch.org/hub/huggingface\_pytorch-transformers/} are employed.

\section{Experiments}
\label{experiment}

\subsection{Experimental Settings}

We conduct our work on NVIDIA Tesla A100 and AMD EPYC 7742 64-Core CPU. During the interpretable language modeling process, BERT-base-Chinese pre-trained model is used, and the original parameters are opted  \footnote{https://huggingface.co/bert-base-chinese}. Additionally, we opt for the rate of $0.3$ for all the dropout layers, a learning rate of 2e-5 for 10-epoch-training of BERT encoder in Fig. \ref{fig:1}a, a learning rate of 1e-5 for 3-epoch-training of the mapping network in Fig. \ref{fig:1}b.

\subsection{Datasets}

\begin{table*}[!t]
\centering
\caption{Statistics information of the used datasets.}
\label{table:dataset}
\begin{tabular}{ccccc}
\hline
Dataset   & TrainSet & TestSet & Dataset Type & Usage                                                           \\ \hline
CLUE      & 2,439    & -       & Article      & \textbf{Initialization} of AKN                                           \\
HybirdSet & 274,039  & 3,162   & Sentence     & \textbf{Training} of correction task                                     \\
SIGHAN15  & 6,526    & 1,100   & Sentence     & \textbf{Evaluation} of correction task                                   \\
%SIGHAN14  & 2,339    & 1,062   & Sentence     & \textbf{Evaluation} of correction task                                   \\
ChnSenti  & 9,600    & 1,089   & Article      & \textbf{Training} and \textbf{Evaluation} of sentiment analysis \\
Weibo100k & 100,000  & 10,000  & Article      & \textbf{Training} and \textbf{Evaluation} of sentiment analysis  \\ \hline
\end{tabular}
\end{table*}

The detailed information of the datasets is exhibited in Table \ref{table:dataset}. \textbf{CLUE} \footnote{https://github.com/CLUEbenchmark/CLUE}, an open-ended, community-driven project, is the most authoritative Chinese natural language understanding benchmark \cite{xu-etal-2020-clue}, where the news dataset is used to initialize associative knowledge network; \textbf{SIGHAN15} is benchmark for traditional Chinese spelling check evaluation \cite{tseng-etal-2015-introduction}, which is widely adopted in simplified Chinese spelling check evaluation by converting to simplified Chinese \cite{cheng-etal-2020-spellgcn,liu-etal-2021-plome}; \textbf{Hybird} is a massive dataset for Chinese spelling correction \cite{wang-etal-2018-hybrid}, which is used the training of the correction methods \cite{wang-etal-2019-confusionset,cheng-etal-2020-spellgcn}; \textbf{Weibo} \footnote{https://github.com/pengming617/bert\_classification} and \textbf{ChnSenti} \footnote{https://github.com/SophonPlus/ChineseNlpCorpus/} sentiment dataset is constructed with the comments from the largest Chinese social community (Sina Weibo) and Chinese hotel reservation websites, which are adopted in the previous work for sentiment classification evaluation \cite{li-etal-2020-enhancing,li-etal-2022-enhancing}. 

\subsection{Comparison Approaches}

We fine-tune our method with standard classifiers of BERT Masked LM and sequence classification for spelling correction and sentiment classification with training of 10 epoch, 1e-5 learning rate and 3 epoch, 1e-5 learning rate.

{\setlength{\parindent}{0cm}
\textbf{SoftMask} is a BERT-based spelling correction method with a soft-mask generator, where the soft-masked strategy is similar to the concept of error detection \cite{zhang-etal-2020-spelling}.
}

{\setlength{\parindent}{0cm}
\textbf{FASPell} conducts the Seq2Seq prediction by incorporating BERT with additional visual and phonology features \cite{hong-etal-2019-faspell}.
}

% SIGHAN 13 14 15
{\setlength{\parindent}{0cm}
\textbf{SpellGCN} incorporates BERT and the graph convolutional network initialized with phonological and visual similarity knowledge for Chinese spelling correction \cite{cheng-etal-2020-spellgcn}.
}

{\setlength{\parindent}{0cm}
\textbf{PLOME} integrates the phonological and visual similarity knowledge into a pre-trained masked language model with a large pre-train corpus consisted of one million Chinese Wikipedia %\footnote{https://zh.wikipedia.org/wiki/}
 pages. And it is the SOTA in previous work \cite{liu-etal-2021-plome}.
}

{\setlength{\parindent}{0cm}
\textbf{HeadFilt} is an adaptable filter for Chinese spell check, which introduce a hierarchical embedding according to the pronunciation similarity and morphological similarity \cite{nguyen2021domain}.
}

{\setlength{\parindent}{0cm}
\textbf{HLG} is a Chinese pre-trained model for word representation by aligning the word-level attention with the word-level distribution with a devised pooling mechanism \cite{li-etal-2020-enhancing}.

\textbf{MWA} introduces a heterogeneous linguistics graph to pre-trained language model. The graph-based structure integrates the linguistics knowledge in the neural network and achieves the SOTA performance in language modeling \cite{li-etal-2022-enhancing}.
}

\textbf{HLG} and \textbf{MWA} is employed on various pre-trained language model, such as vanilla BERT \cite{devlin-etal-2019-bert}, BERT-wwm \cite{cui2021pretrain}, and ERNIE \cite{sun2019ernie}. We use the evaluation results on different pre-trained language models in their original paper.

\subsection{Main Experiments}
\label{main result}

The efficacy of our interpretable language modeling method is evaluated on different tasks, including Chinese spelling correction and sentiment classification. Chinese spelling correction requires the advanced language model for token classification, where every word in the given text is classified into a single class. And the sequence classification is needed in Chinese sentiment classification, where the given text is classified into positive and negative sentiment. The token and sequence classification tasks are able to cover most of the current classification scenario, which enable the efficient demonstration of the language modeling performance of our method.

\subsubsection{Results of Chinese Spelling Correction}

Similar with the past works \cite{cheng-etal-2020-spellgcn, liu-etal-2021-plome}, the correction experiment is employed on word- and sentence-level. Within a more comprehensive perspective, the sentence-level evaluation is wider adopted and more convincing, so we use the same evaluation matrix with the past works \cite{liu-etal-2021-plome, nguyen2021domain}.

As shown in Table \ref{table:correction}, the evaluation on word- and sentence-level composed of different indexes, including detection precision (DP), correction precision (CP), detection recall (DR), correction recall (CR), detection F1 score (DF1) and correction F1 score (CF1). Besides, we assess the influence of the number choosing of $n$ (the size of the coordinates in W2CSpace) and $k$ (the context number in $k$-means clustering algorithm). 
% \cline{3-14}
\renewcommand{\thefootnote}{\fnsymbol{footnote}}
\begin{table*}[!t]
\centering
\caption{Results of Chinese spelling correction.}
\label{table:correction}
\resizebox{\linewidth}{!}{
\renewcommand{\arraystretch}{1}
\begin{tabular}{cccccccccccccc}
\toprule
\multirow{2}{*}{Method}                                                  & \multirow{2}{*}{$k$} & \multicolumn{6}{c}{Word Level}                                                                                              & \multicolumn{6}{c}{Sentence Level}                                                                          \\ \cline{3-14}
                                                                         &                      & DP                    & DR             & DF1                   & CP                    & CR                    & CF1                   & DP             & DR                    & DF1            & CP             & CR                    & CF1            \\ \midrule
FASPell                                                                  & -                    & -                    & -             & -                    & -                    & -                    & -                    & 67.6          & 60.0                 & 63.5          & 66.6          & 59.1                 & 62.6          \\
SoftMask                                                                 & -                    & -                    & -             & -                    & -                    & -                    & -                    & 73.7          & 73.2                 & 73.5          & 66.7          & 66.2                 & 66.4          \\
BERT   & - &92.7    &   85.0       & 88.7   & 96.2   & 81.8     & 88.4  &  76.5    & 78.6   &77.5  & 76.0   & 76.5   &76.3
          \\
PLOME\footnotemark[2]{}                                                              & -                    & \textbf{\emph{94.5}} & 87.4          & \textbf{\emph{90.8}} & \textbf{\emph{97.2}} & \textbf{\emph{84.3}} & \textbf{\emph{90.3}} & 77.4          & \textbf{\emph{81.5}} & 79.4          & 75.3          & \textbf{\emph{79.3}} & 77.2          \\
SpellGCN                                                                 & -                    & 88.9                 & \textbf{87.7} & 88.3                 & 95.7                 & \textbf{83.9}        & 89.4                 & 74.8          & 80.7                 & 77.7          & 72.1          & 77.7                 & 75.9
          \\ 
HeadFilt                                                                 & -                    & -                    & -             & -                    & -                    & -                    & -                    & 84.5          & 71.8                 & 77.6          & 84.2          & 70.2                 & 76.5                               \\ \hline
\multirow{5}{*}{\begin{tabular}[c]{@{}c@{}}W2CSpace\\ $n=50$\end{tabular}}  & $500$                & 90.5                 & 86.2          & 88.3                 & 96.2                 & 82.9                 & 89.0                 & 76.4          & 77.6                 & 77.0          & 75.8          & 75.1                 & 75.5          \\
                                                                         & $1000$               & 90.9                 & 85.8          & 88.2                 & 96.3                 & 82.6                 & 88.9                 & 78.7          & 79.6                 & 79.2          & 78.1          & 76.3                 & 77.4          \\
                                                                         & $1500$               & 91.1                 & 87.1          & 89.0                 & 96.4                 & \textbf{83.9}        & 89.7                 & 78.0          & 79.8                 & 78.9          & 77.4          & 77.0                 & 77.2          \\
                                                                         & $2000$               & 91.7                 & 86.2          & 88.9                 & 96.3                 & 83.0                 & 89.2                 & 79.2          & 79.8                 & 79.5          & 78.5          & 76.6                 & 77.5          \\
                                                                         & $3000$               & \textbf{91.9}        & 86.5          & 89.1       & 96.7                 & 83.6                 & 89.6                 & 78.5          & 80.1                 & 79.3          & 78.0          & 77.7                 & \textbf{77.9} \\ \hline
\multirow{5}{*}{\begin{tabular}[c]{@{}c@{}}W2CSpace\\ $n=100$\end{tabular}} & $500$                & 90.7                 & 86.0          & 88.3                 & 96.3                 & 82.8                 & 89.1                 & 76.6          & 80.5                 & 78.5          & 75.9          & 77.3                 & 76.6          \\
                                                                         & $1000$               & 90.6                 & 86.3          & 88.4                 & 96.2                 & 83.2                 & 89.1                 & 77.2          & 79.8                 & 78.5          & 76.4          & 76.6                 & 76.5          \\
                                                                         & $1500$               & 91.2                 & 86.3          & 88.7                 & 95.8                 & 82.7                 & 88.8                 & \textbf{79.4} & 79.4                 & 79.4          & \textbf{78.8} & 76.6                 & 77.7          \\
                                                                         & $2000$               & 91.4                 & 87.3          & \textbf{89.3}                 & 95.7                 & 83.6                 & 89.2                 & 78.3          & \textbf{80.9}        & \textbf{79.6} & 77.6          & \textbf{77.9}        & 77.8          \\
                                                                         & $3000$               & 90.9                 & 86.4          & 88.6                 & \textbf{96.8}        & 83.7                 & \textbf{89.8}        & 76.9          & 79.7                 & 78.1          & 76.3          & 76.7                 & 76.5          \\ \bottomrule
\end{tabular}
}
\end{table*}

\begin{table}[!t]
\centering
\caption{Results of sentiment classification.}
\label{table:senti}
\begin{tabular}{cccc}
\toprule
Method                                                                   & $k$    & ChnSenti      &  Weibo100K      \\ \midrule
BERT                                                                     & -      & 94.72          & 97.31          \\
\multicolumn{1}{r}{+MWA}                                                 & -      & 95.34          & 98.14          \\
\multicolumn{1}{r}{+HLG}                                                 & -      & 95.83          & 98.17          \\ \hline
BERTwwm                                                                  & -      & 94.38          & 97.36          \\
\multicolumn{1}{r}{+MWA}                                                 & -      & 95.01          & 98.13          \\
\multicolumn{1}{r}{+HLG}                                                 & -      & 95.25          & 98.11          \\ \hline
ERNIE                                                                    & -      & 95.17          & 97.30          \\
\multicolumn{1}{r}{+MWA}                                                 & -      & 95.52          & 98.18          \\
\multicolumn{1}{r}{+HLG}                                                 & -      & 95.83          & 98.22          \\ \hline
\multirow{6}{*}{\begin{tabular}[c]{@{}c@{}}W2CSpace\\ $n=50$\end{tabular}}  & $50$   & 95.70          & 98.22          \\
                                                                         & $100$  & 95.70          & 98.24          \\
                                                                         & $200$  & 95.20          & 98.27          \\
                                                                         & $500$  & 95.45          & 98.30          \\
                                                                         & $800$  & 95.45          & \textbf{98.31}          \\
                                                                         & $1000$ & 95.03          & 98.23          \\ \hline
\multirow{6}{*}{\begin{tabular}[c]{@{}c@{}}W2CSpace\\ $n=100$\end{tabular}} & $50$   & 95.53          & 98.29          \\
                                                                         & $100$  & 94.94          & 98.25          \\
                                                                         & $200$  & 95.62          & 98.27          \\
                                                                         & $500$  & 95.11          & \textbf{98.31}          \\
                                                                         & $800$  & \textbf{95.87} & \textbf{98.31} \\
                                                                         & $1000$ & 95.37          & 98.28          \\ \bottomrule
\end{tabular}
\end{table}

From the correction results in Table \ref{table:correction}, our method outperforms the baselines on both word- and sentence-level. Specifically, at sentence-level, our method respectively advances 0.2 and 0.7 points in DF1 and CF1; at word-level, our method is not able to achieve comparable performance than PLOME \cite{liu-etal-2021-plome}, but advances than the SpellGCN \cite{cheng-etal-2020-spellgcn} with a 0.5 and 0.4 point improvement, and we think the massive training dataset of PLOME significantly enhance the correction performance, where the size of dataset composed of per-train and fine-tune dataset is 600 times larger than ours. 

With respect to the changes of the parameters of W2CSpace, the correction performance are various. W2CSpace cannot achieves best performance on a specific parameter combination, but the overall performance is comparable attractive. Besides, we notice that the W2CSpace with a larger size of the coordinates performs better than the smaller one, where the larger W2CSpace is advanced in 8 indexes. The advantages of enlarging the number $k$ are not obvious, where W2CSpace with $n=100$ and $k=2000$ is the most advanced combination for the correction task. However, W2CSpace with $k=1500$ and $k=3000$ is also a good choice for correction. Generally, we think the introduction of the merge matrix balances the difference of $k$.

\footnotetext[2]{While the other comparison methods are trained on HybirdSet, PLOME additionally pre-trained on a 600 times larger dataset compared with HybirdSet. We uniquely highlight the advanced index of PLOME with \emph{\textbf{bold italic font}.}}

\begin{figure*}[!t]
    \begin{center}
        \includegraphics[width=5in]{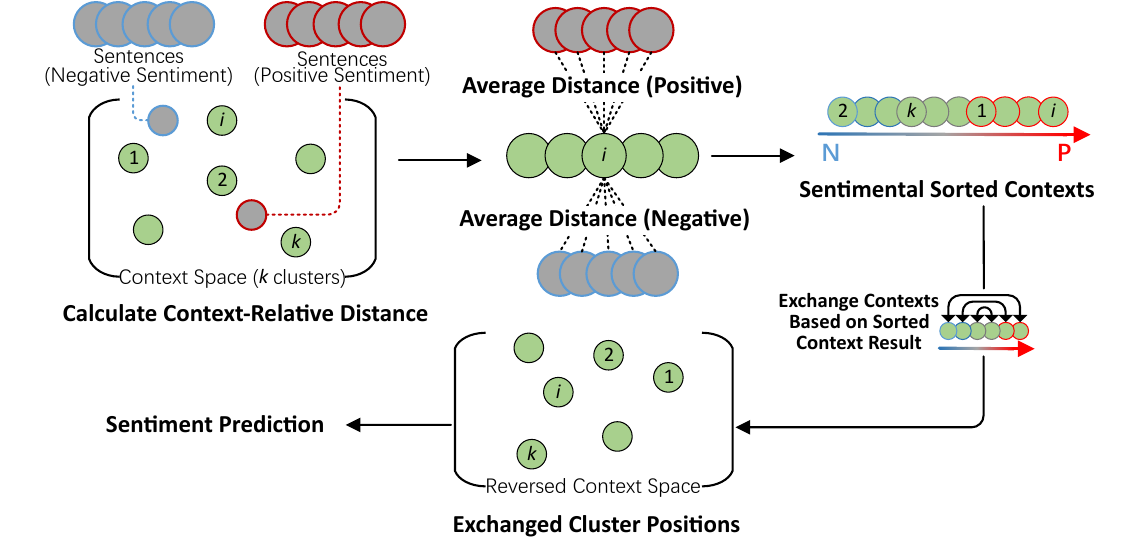}
    \end{center}
    \caption{Interpretable Analysis Procedure. First, we calculate the context-relative distance for both 'Positive' and 'Negative' sentiments. Then we use the average distance between each context and the 'Positive/Negative' sentences to rank the contexts from negative to positive sentiment. By exchanging the positions of the context clusters in the W2CSpace, the contexts are reversed to opposite sentiment based on their average distances. Finally, we apply the reversed context to the sentiment classification task and analyze the changes in the sentiment classification results.}
    \label{fig:interpretable}
\end{figure*}

\subsubsection{Results of Chinese Sentiment Classification}

We evaluate the sentiment classification performance with the classification accuracy. In Table \ref{table:senti}, the results of the sentiment classification are illustrated. Specifically, even the other pre-training models partly perform better than BERT, our method achieves improvements on both ChnSenti and Weibo100K datasets as 0.04 and 0.09 points. 

For the choice of $n$ and $k$ numbers, our method achieves advanced performance with the combination of $n=100$ and $k=800$ for sentiment classification, which is different from the correction experiment. Besides, similar to the tendency in the correction experiment, a larger $n$ number enables a small improvement in performance.

\subsection{Interpretable Analysis}

The interpretable machine learning methods are considered with a transparent decision process \cite{christoph2020Interpretable,verhagen2021two}. However, the method interpretability is hard to define. The rule-based approaches, widely regarded as the interpretable methods, are especially advanced in the controllability that is one of the most significant characteristics of interpretability \cite{lee2017making, Tian_2019_CVPR_Workshops, Tripathy_2020_WACV}. 

The context-relative distance, the key in our interpretable language modeling process, originated from the relativity between the input sentences and the context in W2CSpace. Therefore, we design an analysis focused on the interpretability of context-relative distance by means of Chinese sentiment classification task. The procedure of interpretable analysis is exhibited in Figure \ref{fig:interpretable}. Ideally, the context-relative distance correlates with sentiment, e.g., the shorter distance indicates stronger relativity between the context and the labeled sentiment. If sentiment prediction results change correspondingly after reversing the context space, it convincingly shows that (1) the interpretable knowledge from AKN is integrated into W2CSpace, (2) the feature mapping and context abstraction processes are conducted reasonably, and (3) the distance within W2CSpace is interpretable and associated with the emotion conveyed by the input text. 

The interpretable analysis result on Weibo100K is exhibited in Table \ref{table:control}. OA is the original accuracy, CA is the sentiment classification accuracy after modification, and RA is the reversing accuracy for the sentences that successfully reverse the sentiment labels after modification. Because language modeling mainly relies on context semantics, the predicted sentiments after modification should be reversed compared with the original prediction. From the interpretable results, the values of RA approximate to $100\%$, which indicates that the predicted sentiments are mostly reversed and matches our expectation. \emph{The predictable changes of the accuracy reflects the controllability of our method and the interpretability of W2CSpace.} Besides, from the perspective of model structure, the ideal transparent method enables a completely controllable decision process from input to output. And in our method, even though some parts are still in black-box, but RA reflects the interpretability between the decision processes from W2CSpace to the output and from input to the output. The interpretable contexts are consistent with the linguistics logic in input articles and serves as the agent to cooperate with the articles to realize the controllable process. The value of RA does not directly indicate the interpretability of our method, but \emph{but the more approximate to $1$, the more semantically explainable of W2CSpace and its context}. 

\begin{table}[]
\centering
\caption{Interpretable results.}
\label{table:control}
\begin{tabular}{ccccc}
\toprule
$n$                    & $k$   & OA    & CA   & RA    \\ \midrule
\multirow{3}{*}{$50$}  & $100$ & 98.24 & 3.69  & 96.62  \\
                       & $500$ & 98.30 & 2.96  & 98.12  \\
                       & $1000$ & 98.23 & 1.75  & 99.90  \\ \hline
\multirow{3}{*}{$100$} & $100$ & 98.25 & 1.88 & 99.81 \\
                       & $500$ & 98.31 & 3.79 & 96.78 \\
                       & $1000$ & 98.28 & 2.76 & 98.48     \\ \bottomrule
\end{tabular}
\end{table}

\section{Conclusion}

An interpretable language model is proposed by constructing a word-context-coupled space in this study. Within W2CSpace, (1) the uninterpretable neural representation in BERT is regulated by interpretable associative knowledge relations, (2) an intermediate representation space with reasonable interpretable semantics is designed, (3) an interpretable semantic feature is introduced based on intermediate representation space for downstream classifiers. Obviously, the above strategies bring a strong generalization ability for the interpretable pre-trained language modeling process. Besides, for the potential risk preventing, the interpretable machine learning method is introduced for migrating the adverse affects from the black-box structure. Moreover, in our method, the controllable decision process realize the regulation for the illegal language inputting by controlling the related context, and the strong cooperation between pre-trained models and W2CSpace can protect the parameter privacy from the data stealing.

Nevertheless, W2CSpace is unable to directly handle high-level semantics, including sentences, paragraphs, and so on. Even the word-level language models act as the mainstream methods in NLP, the above limitation should be further considered in the future. Besides, restricted by our knowledge and efforts, the main experiments cannot cover all common tasks and all pre-trained models in NLP. Relatively, the token- and sequence-level classifications have demonstrated attractive experimental performance on most NLP tasks. Next, we also plan to extend W2CSpace to more NLP tasks and find its more specific value.

\section*{Acknowledgements}
This work was supported in part by the National Natural Science Foundation of China (NSFC) under Grant 62272201, and 61872166; in part by the Six Talent Peaks Project of Jiangsu Province under Grant 2019 XYDXX-161.

\bibliography{anthology,custom}
\bibliographystyle{acl_natbib}

\end{document}